\documentclass[letterpaper]{article} 
\usepackage{aaai24}  
\usepackage{times}  
\usepackage{helvet}  
\usepackage{courier}  
\usepackage[hyphens]{url}  
\usepackage{graphicx} 
\usepackage{natbib}  
\usepackage{caption} 
\usepackage{algorithm}
\usepackage{algorithmic}

\usepackage{amsmath}
\usepackage{amsfonts}

\usepackage{color}

\usepackage[table,xcdraw]{xcolor}
\usepackage{newfloat}
\usepackage{listings}
\DeclareCaptionStyle{ruled}{labelfont=normalfont,labelsep=colon,strut=off} 
\floatstyle{ruled}
\newfloat{listing}{tb}{lst}{}
\floatname{listing}{Listing}
\nocopyright

\title{MathAttack: Attacking Large Language Models Towards Math Solving Ability}
\author{
    Zihao Zhou\textsuperscript{\rm 1 2} \quad 
     Qiufeng Wang\textsuperscript{\rm 1 }\quad 
    Mingyu Jin\textsuperscript{\rm 3}\quad Jie Yao\textsuperscript{\rm 1 2}\\ Jianan Ye\textsuperscript{\rm 1 2}\quad Wei Liu\textsuperscript{\rm 4}\quad 
    Wei Wang\textsuperscript{\rm 1}\quad Xiaowei Huang\textsuperscript{\rm 2}\quad Kaizhu Huang\textsuperscript{\rm 5}
}
\affiliations{
    \textsuperscript{\rm 1}Xi’an Jiaotong-liverpool University \quad\quad\quad
    \textsuperscript{\rm 2}University of Liverpool \quad\quad\quad
    \textsuperscript{\rm 3}Northwestern University\\
    \textsuperscript{\rm 4}ShanghaiTech University\quad\quad\quad
    \textsuperscript{\rm 5}Duke Kunshan University

}

\usepackage{bibentry}

\begin{document}

\maketitle

\begin{abstract}
With the boom of Large Language Models (LLMs), the research of solving Math Word Problem (MWP) has recently made great progress. However, there are few studies to examine the security of LLMs in math solving ability. Instead of attacking prompts in the use of LLMs, we propose a \textbf{MathAttack} model to attack MWP samples which are closer to the essence of security in solving math problems.  
Compared to traditional text adversarial attack, it is essential to preserve the mathematical logic of original MWPs during the attacking. To this end, we propose logical entity recognition to identify logical entries which are then frozen. Subsequently, the remaining text are attacked by adopting a word-level attacker. 
Furthermore, we propose a new dataset~\textbf{RobustMath} to evaluate the robustness of LLMs in math solving ability. 
Extensive experiments on our \textbf{RobustMath} and two another math benchmark datasets GSM8K and MultiAirth show that \textbf{MathAttack} could effectively attack the math solving ability of LLMs. In the experiments, we observe that (1) Our adversarial samples from higher-accuracy LLMs are also effective for attacking LLMs with lower accuracy (e.g., transfer from larger to smaller-size LLMs, or from few-shot to zero-shot prompts); (2) Complex MWPs (such as more solving steps, longer text, more numbers) are more vulnerable to attack; (3) We can improve the robustness of LLMs by using our adversarial samples in few-shot prompts. 
Finally, we hope our practice and observation can serve as an important attempt towards enhancing the robustness of LLMs in math solving ability. We will release our code and dataset.

\end{abstract}

\begin{figure}[t]
\centering
\includegraphics[width=0.48\textwidth]{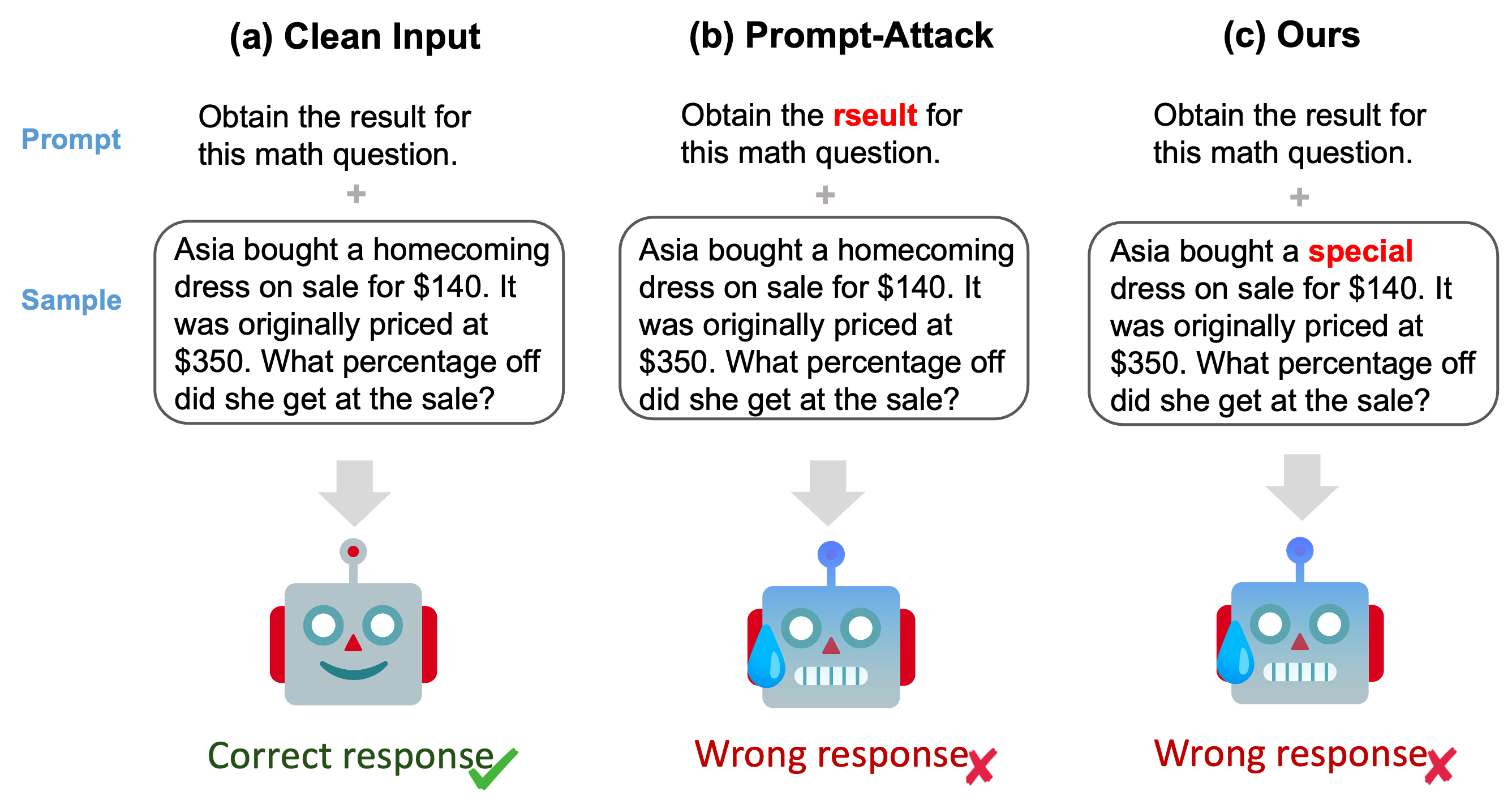} 
\caption{Different input of Large Language Models (LLMs). (a) Clean input, (b) Adversarial sample generated by Prompt-Attack~\cite{zhu2023promptbench,wang2023adversarial}, (c) Adversarial sample generated by our MathAttack.}
\label{introduction}
\end{figure}

\section{Introduction}
Solving Math Word Problem (MWP) aims to infer a final answer from the natural language description of a math problem~\cite{wang2017deep}. With the boom of Large Language Models (LLMs), the research of solving MWP has recently made great progress~\cite{qiao2022reasoning, uesato2022solving, chang2023survey}. Most of them work on prompt engineering to improve math solving ability of LLMs~\cite{wei2022chain,zhou2022least,kojima2022large,chen2022program,fu2022complexity}, and LLMs (e.g., ChatGPT) can provide correct reasoning process and the final answer for simple math word problems. Subsequently, they have been progressively applied in the field of intelligence education~\cite{macina2023mathdial}.  
Therefore, it becomes essential to examine the security of LLMs in math solving ability, but this has not attracted much attention so far. To the best of our knowledge, there are only a few works~\cite{zhu2023promptbench,wang2023adversarial} to evaluate the robustness of LLMs through attacking prompts (Figure~\ref{introduction}(b)). By comparing to prompt-attack, we argue that attacking MWP samples themselves is more direct to reflect the security of LLMs in math solving ability, like Figure~\ref{introduction}(c).


On the other hand, general text adversarial attack has made great progress~\cite{li2018textbugger, li2020bert, ye2022texthoaxer}. This task aims to generate an adversarial text $x'$ that is semantically similar to the original text $x$, while victim model $f$ can correctly classify $x$ but incorrectly classify $x'$~\cite{jin2020bert,xu2020adversarial}. However, it tends to change mathematical logic by directly applying such techniques of general text adversarial attack.
For example, if the word $140$ in Figure~\ref{introduction}(c) is modified to another number, the mathematical logic will be changed and the original ground-truth will be no longer the correct answer. 
Therefore, it is essential to preserve the mathematical logic of MWPs, which makes MWP adversarial attack more challenging.

\begin{figure*}[t]
\centering
\includegraphics[width=0.9\textwidth]{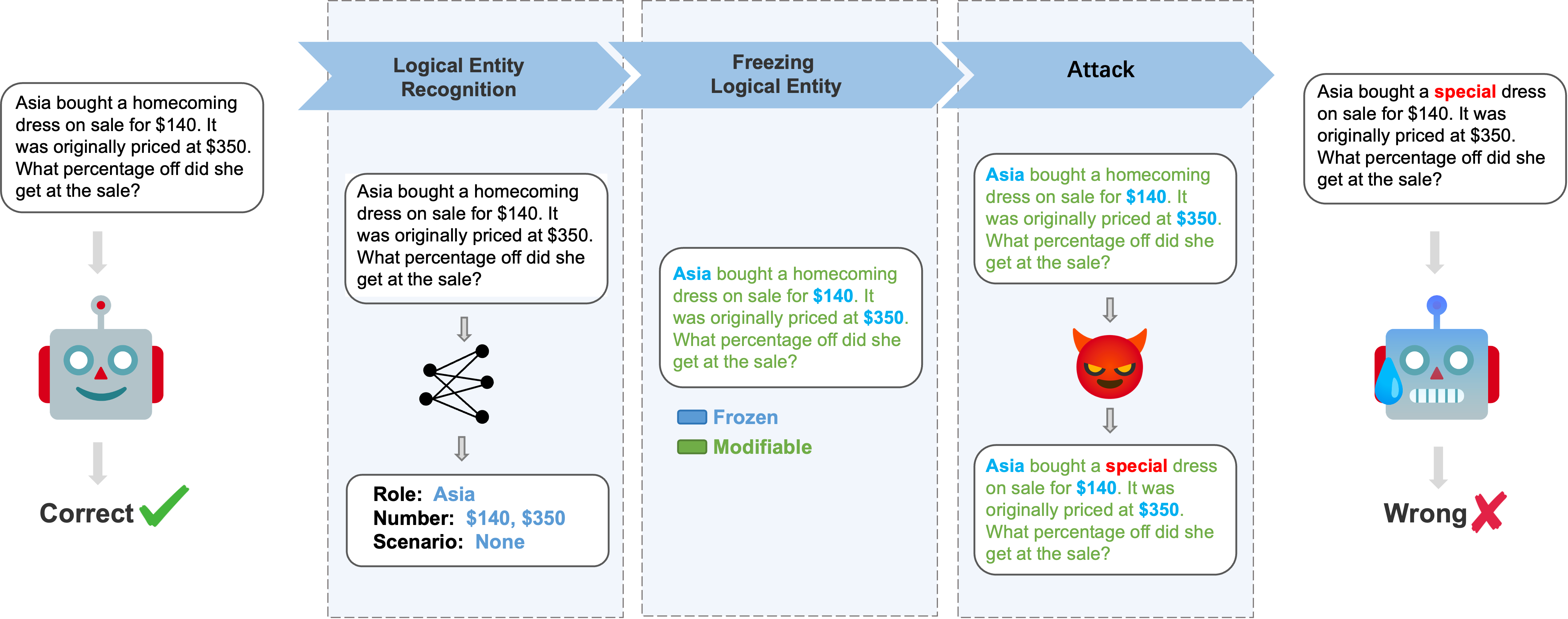} 
\caption{The overview of MathAttack. First, we utilize a NER model to identify logical entities. Then we freeze the logical entities, preventing the attacker from modifying them. Finally, we utilize word-level attacker to attack the LLMs while not changing those frozen logical entities.}
\label{overview}
\end{figure*}

To preserve the mathematical logic of MWPs, we propose \textbf{MathAttack} for attacking the math solving ability of large language models. Figure~\ref{overview} shows an overview of our MathAttack.
We first recognize logical entities, altering these logical entities easily leads to changing the mathematical logic of math word problems.
Then we freeze the logical entities, preventing the attacker from modifying logical entities. Finally we attack the LLMs utilizing word-level attacker~\cite{li2020bert} while not changing those frozen logical words. 
With the help of MathAttack and manual check, we propose a new dataset \textbf{RobustMath}, which consists of 300 high-quality MWP adversarial samples and could measure the robustness of LLMs' math solving ability. 

Extensive experiments on our proposed \textbf{RobustMath} dataset and another two math benchmark datasets GSM8K~\cite{cobbe2021training} and MultiAirth~\cite{roy2015solving} show that our \textbf{MathAttack} could effectively attack the math solving ability of LLMs. 
As far as we know, most works~\cite{zhu2023promptbench,wang2023adversarial} focus the robustness of LLMs in general tasks, there are not any comprehensive study on the security of LLMs in math solving ability. To this end, we conduct a a serious of analysis in the experiments and observe the following three points: (1) Transferability of attacking samples. Adversarial samples generated from higher-accuracy LLMs are also effective for attacking LLMs with lower accuracy (e.g., transfer from larger to smaller-size LLMs, or from few-shot to zero-shot prompts);
(2) Complex MWPs (such as more solving steps, longer text, more numbers) are more vulnerable to attack; (3) We can improve the robustness of LLMs by using our attacking samples in few-shot prompts.

In summary, our contributions are as follows:
\begin{itemize}
\item In this paper, we make a first attempt to attack MWP samples to examine the security of LLMs in math solving ability. 

\item We propose a \textbf{MathAttack} for attacking the math solving ability of large language models, including Logical Entity Recognition, Freezing Logical Entity and text Attack.

\item We propose a new dataset \textbf{RobustMath} by adopting MathAttack and manual check. It consists of 300 high-quality MWP adversarial samples and could measure the robustness of LLMs' math solving ability.

\item Extensive experiments show that MathAttack could effectively attack the math solving ability of LLMs. Through the exhaustive analysis, we obtain three findings for the security of LLMs in math solving ability.
\end{itemize}

\section{Related Work}
\paragraph{MWP Solver}
Recent proposals intend to solve the problem by using sequence or tree generation models. \cite{wang2017deep} presents a sequence-to-sequence (seq2seq) approach to generate the mathematical equation. \cite{xie2019goal} propose a goal-driven tree-structured (GTS) model to generate the equation tree. This sequence-to-tree approach significantly improves the performance over the traditional seq2seq approaches. \cite{zhang2020graph} adopt a graph-to-tree approach to model the quality relations using graph convolutional networks (GCN). Previous studies \cite{patel2021nlp} indicate these MWP solvers rely on shallow heuristics to generate equations. With the boom of Large Language Models (LLMs) and the proposal of chain-of-thought~\cite{wei2022chain}, the math solving ability of the model has recently made great progress. Many research works on prompt engineering to improve math solving ability~\cite{zhou2022least,kojima2022large,chen2022program,fu2022complexity}, they are capable of effortlessly solving simple MWPs, and LLMs are gradually being incorporated in the field of intelligent education~\cite{ji2023systematic, macina2023mathdial}. In this context, examining the security of LLMs in math solving ability becomes essential. In this work, we make a first attempt to examine this security issue by attacking MWP samples.

\paragraph{Large Language Models Attack}
Previous proposals have already tried to evaluate the robustness of large language models \cite{zhuo2023robustness,shi2023large}. \cite{wang2023robustness} makes the first attempt to systematically evaluate the robustness of LLMs by using robust datasets. Recently, some works propose to address this issue by attacking prompts~\cite{wang2023adversarial,zhu2023promptbench}. \cite{wang2023adversarial} introduces the ICL attack based on TextAttack, which aims to manipulate the prompt only without altering the input. \cite{zhu2023promptbench} presents PromptBench, a robustness benchmark specifically designed to evaluate the robustness of LLMs against adversarial prompts. Our work differs from theirs in two main aspects: (1) We specifically focus on attacking the MWP sample itself, which provides a more direct approach and fills the gap of non-prompt attacks on LLMs. (2) Their works target general tasks, lacking a comprehensive analysis of the robustness in math solving ability.
\paragraph{MWP Attack}
For the MWP solvers, previous works generate some MWP adversarial examples by rule-based methods like reordering the problem description \cite{kumar2021adversarial,patel2021nlp}. However, with the development of LLMs, the semantic and logical capabilities of the model have been enhanced, rendering these adversarial examples ineffective. Adversarial MWP sample datasets \textbf{SVAMP} \cite{patel2021nlp} can be solved well by LLMs like ChatGPT. In this paper, we attack MWP samples of LLMs for the first time and propose a new dataset \textbf{RobustMath} to evaluate the robustness of math solving ability of LLMs. It consists of adversarial examples generated by MathAttack, utilizing simple MWPs from GSM8K and MultiAirth as seed data.

\section{Methodology}
\subsection{Problem Formulation}
Suppose we have a text $x$ with $n$ words $x = \left[w_1,w_2, ... ,w_n\right]$ whose ground truth label is $y$. We call $x'$
 an adversarial example when $x'$ can make the victim model $f$ wrong prediction but original correct prediction ($f\left(x\right) = y$), i.e., 
 \begin{equation}
     f\left(x'\right) \neq f\left(x\right).
 \end{equation}
 Compared to traditional text attack, math word problem attack need to preserve the mathematical logical $L$ of text sample $x$, it is defined as:
 \begin{equation}
     L\left(x'\right) = L\left(x\right).
 \end{equation}
The goal of the attack task is to generate an adversarial example $x^*$ among all $x'$. Since text data consists of discrete words whose change can be perceived by humans, we always want the optimized adversarial example $x^*$ to be semantically closest to the original text sample $x$. Thus, the objective function of this task can be defined as follows:
\begin{equation}
    x^* = \mathop{\mathrm{argmax}}\limits_{x'}\mathcal{G}\left(x,x'\right), s.t.f\left(x'\right) \neq f\left(x\right), L\left(x'\right) = L\left(x\right),
\end{equation}
where $\mathcal{G}\left(x,x'\right)$ denotes the semantic similarity between $x$ and $x'$. In this paper, $f$ is the large language model and we follow the black-box setting.

\subsection{The Proposed MathAttack}
\paragraph{Overview} Figure~\ref{overview} shows an overview of MathAttack. We firstly recognize logical entities. Altering these logical entities easily leads to changing the logic of math word problems. Then we freeze the logical entities, preventing the attacker from modifying logical entities. Finally we attack the LLMs utilizing word-level attacker while not changing those frozen logical entities.

\paragraph{Logical Entity Recognition} Logical entities are crucial components that constitute logic in math word problems~\cite{kumar2022practice, li2022semantic}. In order to preserve the logic of a math word problem, it is indispensable to define and identify which entities as logical entities. In this paper, we define the following three types of entities as logical entities. 
(1) \textbf{Role Entity}: It includes person entity (e.g, \textit{Asia} in Figure~\ref{overview}). 
(2) \textbf{Number Entity}: It includes quantity (e.g, \textit{\$140} in Figure~\ref{overview}), cardinal number and ordinal number.
(3) \textbf{Scenario Entity}: It includes time entity and location entity. Altering these environmental factors is easy to change the logic of math word problems too.

Then we employ Named Entity Recognition (NER) model to identify them:
  \begin{equation}
     I_{ro} = NER_{ro}\left(x\right),
  \end{equation}
  \begin{equation}
     I_{num} = NER_{num}\left(x\right),
  \end{equation}
  \begin{equation}
     I_{sce} = NER_{sce}\left(x\right),
 \end{equation}
 where $I_t$ is a word index set if the word belongs to logical entity type $t$. The symbols $ro$, $num$ and $sce$ represent the Role, Number and Scenario Entity respectively. We utilize Spacy \footnote{https://spacy.io/} as our NER model.
\paragraph{Freezing Logical Entity} It tends to break 
the original logic of MWP by altering logical entities during the attack process. To this end,  we freeze all logical entities in order to prohibit attackers from modifying them:
  \begin{equation}
     I_{f} = I_{name}\cup{I_{num}}\cup{I_{sce}},
 \end{equation}
 where $I_f$ denotes the frozen word index set.
\paragraph{Attack} Text attackers are generally classified into three types: char-level, word-level and sentence-level.
In MathAttack, We choose word-level attacker because the char-level attacker can distort the semantic meaning of words (like Figure~\ref{introduction}(b)) and sentence-level attacker are prone to disrupting the mathematical logic of MWP. The attack process of word-level attacker primarily entails two steps: finding vulnerable words and words replacement.

In order to find vulnerable words, it is necessary to determine which words are significant. Specifically, we first sequentially mask all modifiable words to form new sentences. Afterward, we predict each new sentence to get the drop in the probability of the correct answer. The more it drops, the more important the word is. It is defined as:
  \begin{equation}
     x_i^{mask} = \left[w_1,w_2,···,w_{i-1},mask,w_{i+1},...,w_n\right], i\notin I_{f},
 \end{equation}
 \begin{equation}
     a_i = prob\left(f\left(x\right)\right)-prob\left(f\left({x^{mask}_i}\right)\right),
 \end{equation}
where $a_i$ is the important score of $x_i$ and $prob$ is the function to get the probability of the correct answer. After that, we can get the important scores list $a = \left[a_1, a_2, ... , a_n\right]$.  Notice that the length of $a$ is not $n$ because some words are frozen. Finally, we choose the word which has the max important score as the vulnerable word:
  \begin{equation}
     m = argmax\left(a\right)\label{getm},
 \end{equation}
 where $m$ is the index of the vulnerable word and $argmax$ is the function to pop the word which has the most important score and get its index. 

 After finding the vulnerable word, we proceed to locate all synonyms of $w_m$ in order to substitute it:
\begin{equation}
     S = Synonyms\left(w_{m}\right),
 \end{equation}
where $S$ is the synonyms set of $w_m$, we sequentially select a word in $S$ based on the similarity to $w_m$ then substitute $w_m$:
 \begin{equation}
     s' = MaxSim\left(S, w_m\right) \label{gets},
 \end{equation}
  \begin{equation}
     x^s = \left[w_1,w_2,···,w_{m-1},s',w_{m+1},...,w_n\right],
 \end{equation}
 where $x^s$ is the sentence by replacing $w_m$ in $x$ with $s'$.
 $MaxSim$ is the function to pop the word in $S$ that is most similar to $w_m$. Notice that if $S$ is already empty before popping, we go back to Eqn. (\ref{getm}) and repeat the above process. 
 After obtaining $x^s$, we perform different actions based on the following situations, if $f\left(x^s\right) \neq f\left(x\right)$, the final adversarial sample $x^*$ is $x^s$:
  \begin{equation}
  x^* = x^s.
 \end{equation}
 If $f\left(x^s\right) = f\left(x\right)$ and $prob\left(f\left(x^s\right)\right) < prob\left(f\left(x\right)\right)$, we will keep this word change:
 \begin{equation}
  x = x^s.
\end{equation}
Then go back to Eqn. (\ref{gets}) and repeat the above process. If $f\left(x^s\right) = f\left(x\right)$ and $prob\left(f\left(x^s\right)\right) \geq prob\left(f\left(x\right)\right)$, we will abandon this word change then go back to Eqn. (\ref{gets}) and repeat the above process.

In our attacker, we utilize BertAttack~\cite{li2020bert} as our backbone, which utilizes \textit{[mask]} token to mask words and bert embedding to calculate the similarity of words.

\begin{table*}[t]
\centering
\scalebox{0.95}{
\begin{tabular}{cccccccccc}
\hline
          &                                 &       & \textbf{GSM8K} &       &                            &       & \textbf{MultiAirth} &     &     \\ \hline
\textbf{Prompt} &
  \multicolumn{1}{c|}{\textbf{Models}} &
  \textbf{Clean Acc} &
  \textbf{Attack Acc} &
  \textbf{ASR} &
  \multicolumn{1}{c|}{\textbf{Similarity}} &
  \textbf{Clean Acc} &
  \textbf{Attack Acc} &
  \textbf{ASR} &
  \textbf{Similarity} \\ \hline
Zero shot & \multicolumn{1}{c|}{Flan-T5-large} & 18.24 & 2.28           & 87.50 & \multicolumn{1}{c|}{90.55} & 2.00   & 0.00                 & 100.00 & 91.42 \\
          & \multicolumn{1}{c|}{Flan-T5-xl}    & 21.17 & 3.58           & 83.08 & \multicolumn{1}{c|}{92.86} & 7.33  & 0.67                & 90.90 & 95.63 \\
          & \multicolumn{1}{c|}{ChatGLM2}   & 54.40 & 23.78          & 56.29 & \multicolumn{1}{c|}{91.58} & 71.33  & 20.67                & 71.03 & 94.00 \\
          & \multicolumn{1}{c|}{ChatGPT}    & 84.69 & 49.54          & 41.15 & \multicolumn{1}{c|}{89.26} & 98.67  & 60.00                & 39.19 & 91.33 \\ \hline
Few shot  & \multicolumn{1}{c|}{Flan-T5-large} & 22.15 & 10.42          & 52.94 & \multicolumn{1}{c|}{92.92} & 5.33  & 0.67                & 87.5 & 95.66 \\
          & \multicolumn{1}{c|}{Flan-T5-xl}    & 32.35 & 17.59          & 45.45 & \multicolumn{1}{c|}{90.00} & 10.67 & 2.67               & 75.00 & 95.16 \\
          & \multicolumn{1}{c|}{ChatGLM2}   & 64.82 & 22.80          & 64.82 & \multicolumn{1}{c|}{90.71} & 37.33 & 7.33                & 80.36 & 95.75 \\
          & \multicolumn{1}{c|}{ChatGPT}    & 88.27 & 70.68          & 19.93 & \multicolumn{1}{c|}{87.19} & 98.00 & 77.33                & 21.09 & 86.97 \\ \hline
\end{tabular}
}
\caption{Results of attacking against various large language models.}
\label{main result}
\end{table*}

\section{Experiments}
\subsection{Experimental Setting}
\paragraph{Victim Models}
We choose four mainstream large language models as our victim models.
\begin{itemize}
\item 
\textbf{Flan-T5-large}~\cite{chung2022scaling}: Flan-T5-large is a derivative of the Text-to-Text Transfer Transformer (T5) model, developed by Google. It has 760M parameters.
\item 
\textbf{Flan-T5-xl}~\cite{chung2022scaling}: Flan-T5-xl is a large version of Flan-T5 than Flan-T5-large, developed by Google. It has 3B parameters.
\item 
\textbf{ChatGLM2}~\cite{du2022glm}: ChatGLM2 is the second-generation version of the open-source bilingual (Chinese-English) chat model ChatGLM, developed by Tsinghua University. It has 6B parameters.
\item
\textbf{ChatGPT}~\cite{chatgpt}: Developed by OpenAI, ChatGPT is a large language model trained to generate human-like text. It uses the GPT-3 architecture and has been fine-tuned for more interactive and conversational tasks. In detail, we use the gpt-3.5-turbo API.
\end{itemize}
We set the temperature = 0 to stabilize the output of LLMs.
When attacking victim models, we not only attack them with zero-shot prompt but also few-shot prompt. Specifically, we employ four MWP samples as shots and provide Chain-of-Thought (CoT) \cite{wei2022chain} annotations. This few-shot prompt serves as a method to enhance the math solving ability of LLMs. Similar with other prompts, they are not changed during the attack process.
\paragraph{Datasets}
Two math word problems benchmark datasets \textbf{GSM8K} \cite{cobbe2021training} and \textbf{MultiArith} \cite{roy2015solving} are adopted in the experiments. However, we only select the subsets for the following considerations by following the previous work~\cite{zhu2023promptbench}: (1) we focus on simple MWPs, because hard samples have very lower accuracy not necessary to attack. (2) Owing to the extensive computational requirements of generating single adversarial sample, which necessitates iterating over the entire dataset 100 times on average. 
Finally, for GSM8K, we firstly remove those hard samples labelled by more than three solving steps, then randomly select half of those remained simple MWPs and obtain 307 MWP samples.
For MultiAirth, all MWPs are simple thus we randomly select 150 MWPs similar with the previous work~\cite{zhu2023promptbench}.

\paragraph{Metrics}
Given a dataset $D$ with $N$ data instance $x$ and label $y$, victim model $f$, an adversarial attack method $A$ that generates adversarial examples $A(x)$, we adopt following four metrics:
\begin{itemize}
\item 
\textbf{Clean Acc}: The accuracy before attacking. Clean Acc = $\frac{\sum_{(x,y)\in D} \mathbb{I} [f(x) = y]}{N}$.
\item 
\textbf{Attack Acc}: The accuracy after attacking.
 Attack Acc = $\frac{\sum_{(x,y)\in D} \mathbb{I} [f(x)=y \cap f(A(x))=y]}{N}$.
\item
\textbf{Attack Success Rate (ASR)} \cite{wang2023adversarial}: 
The rate of samples is successfully attacked. ASR = $\frac{\sum_{(x,y)\in D} \mathbb{I} [f(A(x)) \neq y]}{\sum_{(x,y)\in D} \mathbb{I} [f(x) = y]}$.
\item 
\textbf{Similarity}: The average semantic similarity between the adversarial sample and the original sample. 
We use Universal Sentence Encoder \cite{cer2018universal}  to measure semantic similarity.
\end{itemize}
To ensure the correctness in the experiments, we check each adversarial sample manually, and consider adversarial examples which are changed mathematical logic as unsuccessful attacks.

\begin{figure}[t]
\centering
\includegraphics[width=0.45\textwidth]{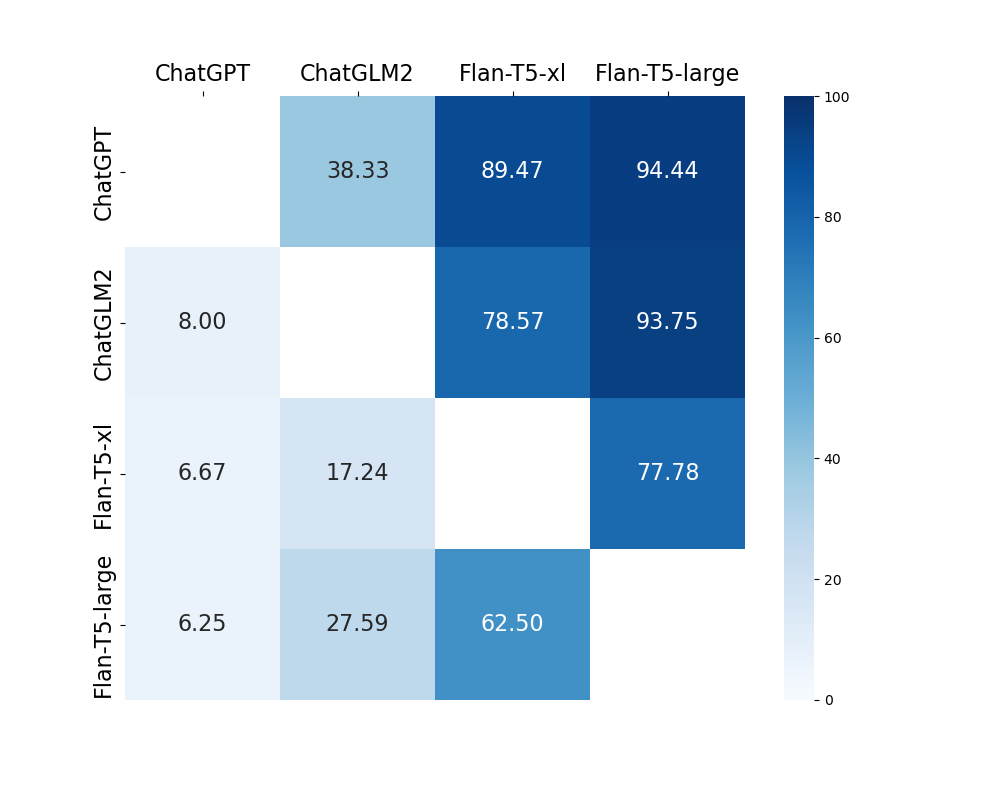} 
\caption{
Transfer Success Rate (TSR) of Y-axis models to X-axis models. The generated adversarial samples of larger-size models can attack smaller-size models.}
\label{trans_m}
\end{figure}

\begin{figure}[t]
\centering
\includegraphics[width=0.32\textwidth]{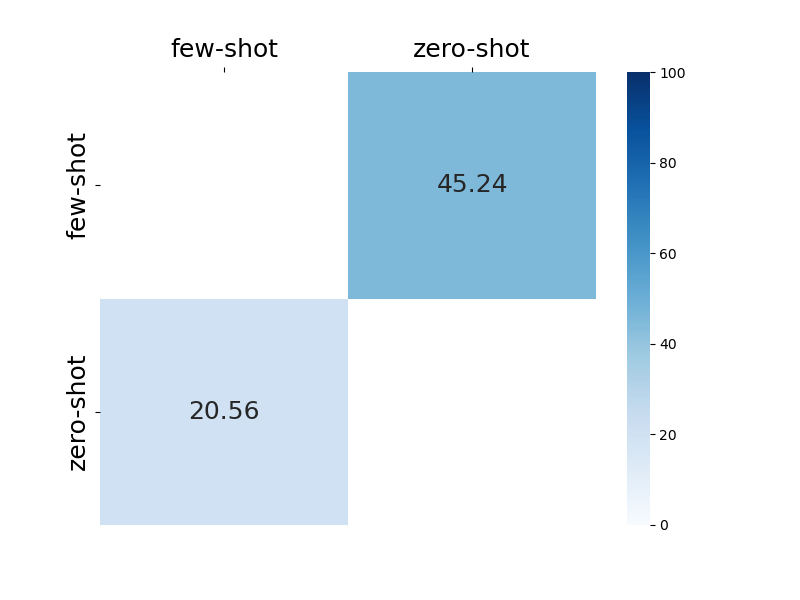} 
\caption{Transfer Success Rate (TSR) of Y-axis prompt to X-axis prompt. The generated adversarial samples of model with few-shot can attack model with zero-shot.}
\label{trans_p}
\end{figure}

\subsection{Main results}
As shown in Table \ref{main result}, our approach can effectively attack the math solving ability of large language models. For LLMs with zero-shot, we could get the high ASR, even for ChatGPT, it could achieve an average of 40\% on GSM8K (41.15\%) and MultiAirth (39.19\%). The average Similarity is large than 90\%, indicating that we can successfully generate adversarial samples with high similarity and do not alter mathematical logic. 

Comparing different LLMs, we can observe that more powerful LLMs (i.e., higher Clean Acc) are more difficult to attack (i.e., lower ASR). 
For Flan-T5-large and Flan-T5-xl, their robustness in math solving ability is poor, as even a slight disturbance can cause them to predict incorrectly. For ChatGLM2 and ChatGPT, their robustness is noticeably stronger, as our method fails to attack them on some MWP samples.

Furthermore, comparing zero-shot and few-shot, we can see that employing few-shot could enhance the math solving ability of the LLMs and also make them more robust, leading to a lower ASR. For models with stronger in-context ability, the enhancement becomes larger. Like ChatGPT, the Attack Success Rate could decrease from 41.15\% to 19.93\%. 
However, we find ChatGLM2 exhibits poor in-context ability which leads math solving ability as well as robustness does not improve with the few-shot prompt.

\begin{figure*}[t]
\centering
\includegraphics[width=0.95\textwidth]{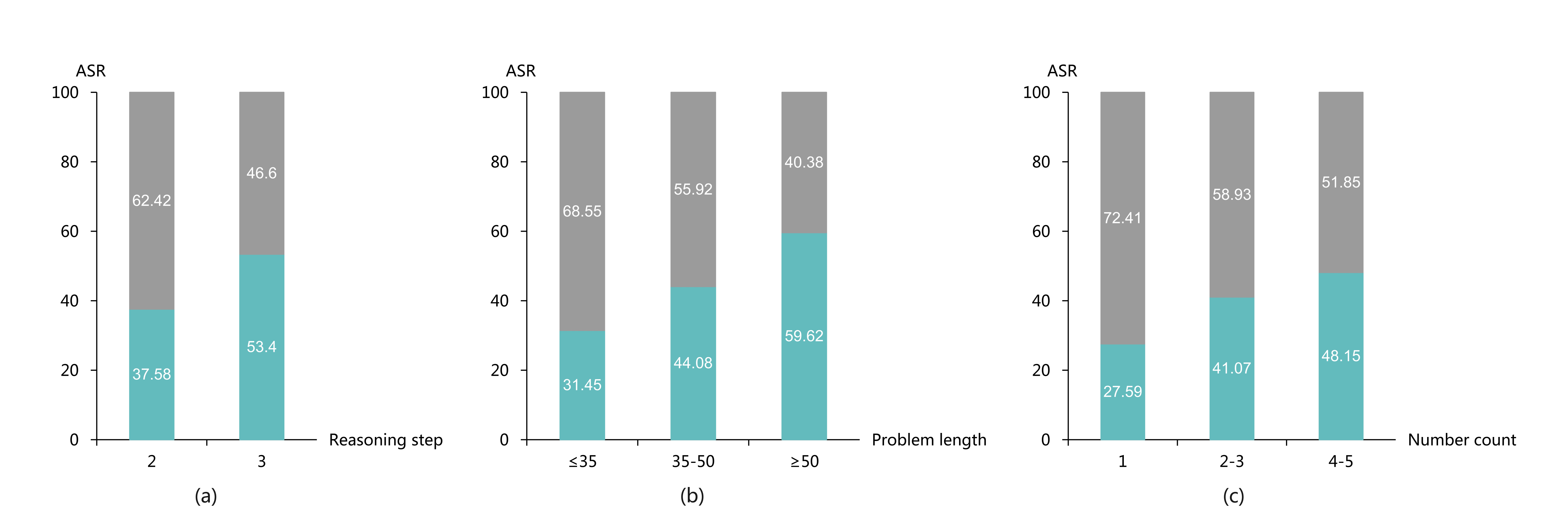}
\caption{Analysis which MWPs are easier to attack. \textbf{(a)} shows the effect of answer reasoning steps on the ASR. \textbf{(b)} shows the effect of problem length on the ASR. \textbf{(c)} shows the effect of numbers' count in MWP on the ASR.}
\label{ana_adv}
\end{figure*}

\subsection{Fine-grained Analysis}
\paragraph{Transferability}
To test the transferability of the generated adversarial samples, we take adversarial samples of model $A$ to attack other models $B$. Specifically, we select the samples that $B$ can correctly predict as the experimental samples. Subsequently, we provide $B$ with adversarial samples generated by attacking $A$ on experimental samples. We examine if these adversarial samples can successfully attack model $B$. Here, we propose the metric: Transfer Success Rate (\textbf{TSR}), if an adversarial sample of $A$ can successfully attack model $B$ then it means transfer success.

In Figure~\ref{trans_m}, we show the TSR between Y-axis (i.e., $A$ model) and X-axis (i.e., $B$ model) models, and we can observe that the adversarial samples of larger-size models can attack smaller-size models too. ChatGPT could get 94.44\% TSR to Flan-T5-large and 89.47\% TSR to Flan-T5-xl. Specifically, we find that the TSR will increase when the math solving ability between models grows wider. As shown in Figure~\ref{trans_m}, we can see that the adversarial samples of smaller-size models can not attack larger-size models. Flan-T5-large and Flan-T5-xl both show low TSR (6.25\% and 6.67\%) on ChatGPT.

In order to see the transferability performance between zero-shot and few-shot, we conducted the same experiment on ChatGPT.
As shown in Figure~\ref{trans_p}, the ChatGPT with few-shot can achieve 45.24\% TSR to ChatGPT with zero-shot however the reverse is only 20.56\%. It indicates that the adversarial samples of LLM with few-shot can attack that with zero-shot. And adversarial samples of LLM with zero-shot can not transfer to that with few-shot.

\paragraph{Analysis on MWPs}
To know which MWPs are easier to attack, we investigate the effects of MWPs
reasoning steps, problem length and number count on ASR. As shown in Figure~\ref{ana_adv}: (a) with the increase of reasoning steps of ground truth, we can observe that the ASR will increase when the reasoning steps from 2 to 3. Reasoning steps of ground truth can be regarded as a metric to measure the difficulty of MWP. Difficult MWPs are easier to attack.
(b) with the increase in problem length, we can observe a gradual increase in the ASR as the length of the math word problems become longer.
(c) with the increase in the quantity of numbers in MWPs, we can observe a gradual increase in ASR as the number counts become more. All the above three factors can be used to measure the complexity of an MWP \cite{fu2022complexity}, therefore, we can draw a conclusion that more complex MWPs are easier to attack. 

\begin{table}[]
\centering
\scalebox{0.85}{
\begin{tabular}{|c|c|c|c|c|}
\hline
                    & \textbf{Clean Acc} & \textbf{Attack Acc} & \textbf{ASR} & \textbf{Similarity} \\ \hline
\textbf{GSM8K}      & 87.95              & 75.57               & 14.07        & 88.33               \\ \hline
\textbf{MultiAirth} & 98.00              & 82.00               & 16.33        & 88.15               \\ \hline
\end{tabular}
}
\caption{Results of attacking against large language models with adversarial samples prompt.}
\label{adv_as_pro}
\end{table}

\begin{table*}[t]
\centering
\scalebox{0.67}{
\begin{tabular}{|l|l|}
\hline
\rowcolor[HTML]{DAE8FC} 
\multicolumn{1}{|c|}{\cellcolor[HTML]{DAE8FC}\textbf{Orginal Sample}} &
  \multicolumn{1}{c|}{\cellcolor[HTML]{DAE8FC}\textbf{Adversarial Sample}} \\ \hline
\rowcolor[HTML]{FFFFFF} 
\begin{tabular}[c]{@{}l@{}}\\A \textcolor{green}{class} of 50 students has various hobbies. 10 like to bake, 5 like to play basketball, and\\ the rest like to either play video games or play music. How many like to play video games \\ if the number that like to play music is twice the number that prefer playing basketball?\\ \\ \end{tabular} &
  \begin{tabular}[c]{@{}l@{}}\\A \textcolor{red}{group} of 50 students has various hobbies. 10 like to bake, 5 like to play basketball, and \\ the rest like to either play video games or play music. How many like to play video games \\ if the number that like to play music is twice the number that prefer playing basketball?\\ \\\end{tabular} \\ \hline
\begin{tabular}[c]{@{}l@{}}\\ \textbf{Solution}: \\ Let's assume the number of students who like to play video games is "x". \\ According to the given information, the number of students who like to play music is \\ twice the number that prefer playing basketball. So, the number of students who like to\\ play music is 2 * 5 = 10. Now, we can calculate the total number of students who like to\\ play video games or play music. \\ This can be done by adding the number of students who like to play video games (x) and\\ the number of students who like to play music (10). Total number of students who like to \\ play video games or play music = x + 10. We also know that the total number of students\\  in the class is 50. So, we can set up an equation:\\ 10 (students who like to bake) + 5 (students who like to play basketball) + x (students who\\ like to play video games) + 10 (students who like to play music) = 50 \\ Simplifying the equation: 10 + 5 + x + 10 = 50.   25 + x = 50.   x = 50 - 25.  x = 25.    \\ Therefore, the number of students who like to play video games is 25. (\textcolor{green}{\textbf{Correct!}})\\ \\ \end{tabular} &
  \cellcolor[HTML]{FFFFFF}\begin{tabular}[c]{@{}l@{}}\textbf{Solution}\\  Let's assume the number of students who like to play video games is "x". \\ According to the given information, the number of students who like to play music is\\ twice the number that prefer playing basketball. So, the number of students who like to\\ play music is 2 * 5 = 10. Now, we can calculate the total number of students who like to\\ play video games or play music. \\ This would be the sum of the number of students who like to play video games (x) and \\ the number of students who like to play music (10). Total number of students who like to\\ play video games or play music = x + 10. Since the total number of students in the group \\ is 50, we can set up the equation:\\ \textcolor{red}{x + 10 = 50.}  \\ Now, let's solve for x:  x = 50 - 10.  x = 40. \\ Therefore, the number of students who like to play video games is 40. (\textcolor{red}{\textbf{Wrong!}})\end{tabular} \\ \hline
\end{tabular}
}
\caption{A real case predicted by ChatGPT on original MWP (left) and its adversarial sample (right).}
\label{case}
\end{table*}

\begin{table}[]
\scalebox{0.8}{
\begin{tabular}{ccccc}
\hline
                                            & \multicolumn{2}{c}{\textbf{Zero-shot}} & \multicolumn{2}{c}{\textbf{Few-shot}} \\ \hline
\multicolumn{1}{c|}{} &
  \textbf{Original} &
  \multicolumn{1}{c|}{\textbf{RobustMath}} &
  \textbf{Original} &
  \textbf{RobustMath} \\ \hline
\multicolumn{1}{c|}{\textbf{Flan-T5-large}} & 10.75   & \multicolumn{1}{c|}{4.67}    & 13.08             & 10.67             \\ \hline
\multicolumn{1}{c|}{\textbf{Flan-T5-xl}}    & 17.76   & \multicolumn{1}{c|}{6.00}    & 26.17             & 20.33             \\ \hline
\multicolumn{1}{c|}{\begin{tabular}[c]{@{}c@{}}\textbf{Flan-T5-xl-F}\end{tabular}} &
  16.36 &
  \multicolumn{1}{c|}{12.33} &
  10.19 &
  9.33 \\ \hline
\multicolumn{1}{c|}{\textbf{ChatGLM2}}      & 47.08   & \multicolumn{1}{c|}{36.67}   & 54.67             & 33.67             \\ \hline
\end{tabular}}
\caption{Accuracy of large language models on RobustMath and its original samples set. \textbf{Flan-T5-xl-F} is the finetune model on 200k MWP data \cite{fu2023specializing}.}
\label{result of robust}
\end{table}

\begin{figure}[t]
\centering
\includegraphics[width=0.47\textwidth]{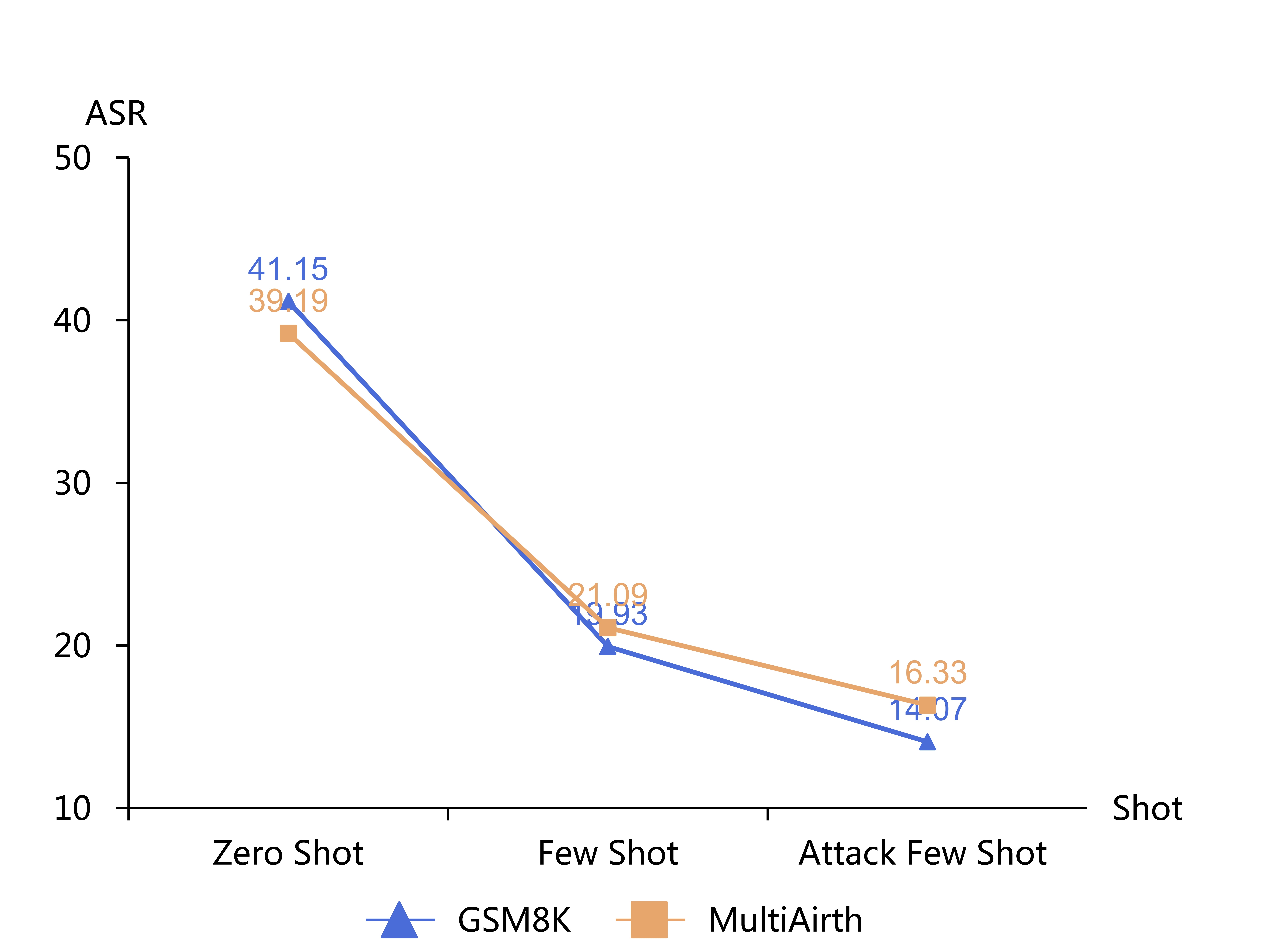} 
\caption{Trend of ASR after utilizing different prompts, Attack Few Shot means replacing the MWP samples of Few Shot to their corresponding adversarial samples.}
\label{trend of asr}
\end{figure}

\paragraph{Using Attacking Samples as Prompts}
In the few-shot prompts, we replace normal MWP examples by corresponding adversarial examples generated by our MathAttack but with correct labels, and observe their impact on the math solving ability and robustness of the LLMs.
As shown in Table~\ref{adv_as_pro}, we can see the Clean ACC still maintains a high level of accuracy (87.95\% on GSM8K and 98.00\% on MultiAirth), because the adversarial examples generated by our MathAttack exhibit high similarity to the original samples. By comparing the Attack Acc and ASR in Table~\ref{main result}, it is surprised to find the use of adversarial examples in the few-shot prompts can enhance the robustness of LLMs (i.e., much lower ASR by comparing to the normal results in Table~\ref{main result}).
When we use adversarial examples in the prompt, the LLM could see these examples that are disturbed but still able to predict correctly, therefore they will not affected by some small disturbances when predict. 
Figure~\ref{trend of asr} provides a more intuitive visualization, demonstrating that the robustness of LLMs utilizing few-shot prompt can be significantly improved by comparing to zero-shot prompt. When employing adversarial examples as few-shot prompt, 
it will further strengthen their robustness and the ASR of large language models further decrease. This observation motivates us to enhance the robustness of large language models without compromising their math solving ability by employing the adversarial examples generated by MathAttack as few-shot prompt. 

\subsection{Case Study}
Table~\ref{case} reports a real case predicted by ChatGPT on the original MWP and its adversarial sample generated by MathAttack. We can find that the adversarial sample generated by MathAttack is similar to the original sample with few changes. For the original sample, ChatGPT can give the correct reasoning process step by step and finally get the correct answer \textit{25}. But when MathAttack simply changed \textit{class} in the original sample to \textit{group}, ChatGPT can come up with the wrong reasoning process and get the wrong equation (\textit{x+10=50}), end up with the wrong answer \textit{40}. We show more real cases predicted by ChatGPT on adversarial samples in \textbf{Appendix A}. These cases show that the robustness of LLMs in math solving ability still needs to be strengthened.

\subsection{New MWP Dataset RobustMath}
Using the transferability of adversarial samples, we attack ChatGPT by MathAttack to build our \textbf{RobustMath} dataset. Specifically, we first utilize GSM8K and MultiAirth as our seed data then attack ChatGPT to generate adversarial samples. After that, in order to ensure the high quality of RobustMath, we manually check each adversarial sample and filter out samples which change the mathematical logic. Ultimately, our RobustMath has 300 high-quality adversarial samples that can be used to measure the robustness of large language models' math solving ability. RobustMath is available at this link, and \textbf{Appendix B} shows some samples of RobustMath.

To verify the effectiveness of our RobustMath, we evaluate large language models on RobustMath. In addition to the models mentioned above, we also evaluate large language model that is fine-tuned on MWP datasets. Specifically, we follow \cite{fu2023specializing} to finetune Flan-T5-xl with 200k MWP data. In Table~\ref{result of robust}, we observe that the performance of the LLMs on RobustMath is significantly worse compared to the performance on its original samples. This indicates that our RobustMath can effectively measure the robustness of the model's math solving ability. When examining the performance of models with zero-shot performance, we can see that as the model's capability increases, its performance on RobustMath also increases. However, it still does not exceed 37.00\%. Moreover, after finetuning, the performance of Flan-T5-xl increases from 6.00\% to 12.33\%. It indicates that finetuning on specific data could help improve the robustness of models. Observing the performance of models with few-shot prompt, we find that models with a strong in-context ability such as Flan-T5-large and Flan-T5-xl can effectively enhance their performance on RobustMath. In contrast, ChatGLM2 and finetuned Flan-T5-xl which have weaker in-context ability do not exhibit significant improvements on both the original samples set and RobustMath under few-shot prompt.


\section{Conclusion and Future Work}
In this paper, we make a first attempt to attack MWP samples to examine the security of LLMs in math solving ability. To preserve the mathematical logic of MWPs, we propose a \textbf{MathAttack} model with a logical entity recognition block. 
Extensive experiments show that MathAttack could effectively attack the math solving ability. Through the comprehensive experimental analysis, we have three significant findings: (1) Transferability of attacking samples (2) Complex MWPs (such as more solving steps, longer text, more numbers) are more vulnerable to attack, and (3) Attacking samples used in few-shot prompts can improve robustness of LLMs. Furthermore, we propose a new dataset \textbf{RobustMath} by utilizing MathAttack and manual check, which consists of high-quality MWP Adversarial samples and could measure the robustness of LLMs' math solving ability. We hope our practice and observations can serve as an important attempt to enhance the robustness of LLMs in math solving ability. In the future, we will explore methods such as instruction learning or reinforcement learning to enhance the robustness of the models. As large language models are increasingly being applied in the field of intelligence education, the importance of improving their robustness becomes more significant.

\bibliography{ref}

\section{Appendix}
\subsection{A}
Table~\ref{predict} shows three cases predicted by ChatGPT on original samples and their adversarial samples. 
\subsection{B}
Table~\ref{robustM} shows eight samples from our proposed \textbf{RobustMath} dataset and their corresponding adversarial samples.

\begin{table*}[t]
\centering
\scalebox{0.8}{
\begin{tabular}{|l|l|}
\hline
\multicolumn{1}{|c|}{\textbf{Original Sample}} &
  \multicolumn{1}{c|}{\textbf{Adversarial Sample}} \\ \hline
\begin{tabular}[c]{@{}l@{}}\textbf{MWP1:}\\\\Brett is 14 years old \textcolor{green}{.} in four years his sister Angela will be \\ three times as old as he is now . how old is Angela right now ?\\ \\ \end{tabular} &
  \begin{tabular}[c]{@{}l@{}}\\\\Brett is 14 years old \textcolor{red}{;} in four years his sister Angela will be \\ three times as old as he is now . how old is Angela right now ? \\ \\ \end{tabular} \\ \hline
\begin{tabular}[c]{@{}l@{}}\\Solution:\\ Let's think step by step. In four years, Brett will be 14 + 4\\  = 18 years old. Angela will be three times as old as Brett \\ is now, so Angela will be 3 * 14 = 42 years old in four years. \\ Therefore, Angela is currently 42 - 4 = 38 years old. (\textcolor{green}{Correct!}) \\ \\ \end{tabular} &
  \begin{tabular}[c]{@{}l@{}}Solution:\\ Let's think step by step. In four years, Brett will be 14 + 4 \\ = 18 years old. Angela will be three times as old as Brett \\ is now, \textcolor{red}{so Angela is currently 18 / 3 = 6 years old}. (\textcolor{red}{Wrong!})\end{tabular} \\ \hline
\begin{tabular}[c]{@{}l@{}} \textbf{MWP2:}\\ \\At the \textcolor{green}{arcade} Cody won 49 tickets . if he spent 25 tickets on \\ a beanie and later won 6 more tickets , how many would \\ he have ? \\ \\\end{tabular} &
  \begin{tabular}[c]{@{}l@{}}\\\\At the \textcolor{red}{game} Cody won 49 tickets . if he spent 25 tickets on\\ a beanie and later won 6 more tickets , how many would\\ he have ? \\ \\ \end{tabular} \\ \hline
\begin{tabular}[c]{@{}l@{}}Solution:\\ To find out how many tickets Cody would have after \\ spending 25 tickets on a beanie and winning 6 more tickets, \\ we can subtract the tickets spent and add the tickets won to\\ the initial number of tickets.\\ \\ Initial number of tickets = 49\\ Tickets spent on a beanie = 25\\ Tickets won = 6\\ Cody would have (49 - 25 + 6) = 30 tickets. (\textcolor{green}{Correct!})\end{tabular} &
  \begin{tabular}[c]{@{}l@{}}\\Solution:\\ To find out how many tickets Cody would have after \\ spending 25 tickets on a beanie and winning 6 more tickets, \\ we can subtract the spent tickets and add the won tickets to \\ his initial total of 49 tickets.\\ \\ Initial total: 49 tickets\\ Spent on a beanie: 25 tickets\\ Won later: 6 tickets\\ Total tickets after spending and winning: 49 - 25 + 6 = \textcolor{red}{30 + 6} = 36\\ Therefore, Cody would have 36 tickets. (\textcolor{red}{Wrong!})\\ \\ \end{tabular} \\ \hline
\begin{tabular}[c]{@{}l@{}} \textbf{MWP3:}\\\\Chloe \textcolor{green}{picked} 48 carrots from her garden . if she threw out \\ 45 of them and then picked 42 more the next day , how \\ many carrots would she have total ? \\ \\ \end{tabular} &
  \begin{tabular}[c]{@{}l@{}}\\\\Chloe \textcolor{red}{selected} 48 carrots from her garden . if she threw out \\ 45 of them and then picked 42 more the next day , how \\ many carrots would she have total ? \\ \\\end{tabular} \\ \hline
\begin{tabular}[c]{@{}l@{}}Solution:\\ To find the total number of carrots Chloe would have,we \\ need to subtract the carrots she threw out and then add \\ the carrots she picked the next day. Total number of \\ carrots = (48 - 45) + 42\\ \\ Simplifying the equation, we get:\\ Total number of carrots = 3 + 42\\ Therefore, the answer is 45. (\textcolor{green}{Correct!})\end{tabular} &
  \begin{tabular}[c]{@{}l@{}}\\Solution:\\To find the total number of carrots Chloe would have, we \\ need to subtract the carrots she threw out and then add \\ the carrots she picked the next day.\\ Carrots picked from the garden: 48\\ Carrots thrown out: 45\\ Carrots picked the next day: 42\\ Total carrots = Carrots picked from the garden -\\ Carrots thrown out + Carrots picked the next day\\ Total carrots = 48 - 45 + 42\\ \\Simplifying the equation:\\ Total carrots = 3 + 42    Total carrots = 45.  Therefore, Chloe would \\ have a total of 45 carrots.\\ The answer is \textcolor{red}{45 + 42 = 87}. Chloe would have a total of 87 carrots. \\ (\textcolor{red}{Wrong!})\end{tabular}  \\ \hline
\end{tabular}
}
\caption{Three cases predicted by ChatGPT on original MWPs (left) and their adversarial samples (right).}
\label{predict}
\end{table*}

\begin{table*}[]
\centering
\scalebox{0.9}{
\begin{tabular}{|l|l|}
\hline
\multicolumn{1}{|c|}{\textbf{Original Sample}} &
  \multicolumn{1}{c|}{\textbf{RobustMath}} \\ \hline
\begin{tabular}[c]{@{}l@{}}\\\\After transferring to a new \textcolor{green}{school}, \\ Amy made 20 more friends than \\ Lily . If Lily made 50 friends , how \\ many friends do Lily and Amy have \\ together ?\\\\\end{tabular} &
  \begin{tabular}[c]{@{}l@{}}\\\\After transferring to a new \textcolor{red}{class )} \\ Amy made 20 more friends than \\ Lily . if Lily made 50 friends , how \\ many friends do Lily and Amy have \\ together ?\\\\\end{tabular} \\ \hline
\begin{tabular}[c]{@{}l@{}}\\\\\textcolor{green}{Mrs} . Sherman made a dozen bread \\ rolls for breakfast . After feeding \\ her 6 children with one each , she \\ broke each of the remaining rolls \\ into 8 pieces and fed them to the \\ chickens .  How many pieces of rolls \\ did she feed to the chickens ?\\\\\end{tabular} &
  \begin{tabular}[c]{@{}l@{}}\\\\\textcolor{red}{.} . Sherman made a dozen bread \\ rolls for breakfast . After feeding \\ her 6 children with one each , she \\ broke each of the remaining rolls \\ into 8 pieces and fed them to the \\ chickens . How many pieces of rolls \\ did she feed to the chickens ?\\\\\end{tabular} \\ \hline
\begin{tabular}[c]{@{}l@{}}\\\\A \textcolor{green}{teacher} had 38 worksheets to grade. \\if she graded 4 , but then another 
15 \\were turned in , how many 
worksheets \\would she have to 
grade ? \\\\\end{tabular} &
  \begin{tabular}[c]{@{}l@{}}\\\\A \textcolor{red}{person} had 38 worksheets to grade. \\if she graded 4 , but then another 
15 \\were turned in , how many 
worksheets \\would she have to 
grade ? \\\\\end{tabular} \\ \hline
\begin{tabular}[c]{@{}l@{}}\\\\Stetson \textcolor{green}{made} a bet with Alec that \\ he would give up \$ 10 for each \\ orange he eats . While at the farm, \\ Stetson ate 2 / 5 of the oranges \\ they picked . If they picked 60 \\ oranges , calculate the total amount \\ of money Stetson gave up ?\\\\\end{tabular} &
  \begin{tabular}[c]{@{}l@{}}\\\\Stetson \textcolor{red}{did} a bet with Alec that \\ he would give up \$ 10 for each \\ orange he eats . While at the farm, \\ Stetson ate 2 / 5 of the oranges \\ they picked . If they picked 60 \\ oranges , calculate the total amount \\ of money Stetson gave up ?\\\\\end{tabular} \\ \hline
\begin{tabular}[c]{@{}l@{}}\\\\
Paige had 8 songs on her mp3 \textcolor{green}{player .} \\If she deleted 5 \textcolor{green}{old} songs from it and \\then added 30 new songs, how many \\songs does she have on her mp3 player ?
\\\\\end{tabular} &
  \begin{tabular}[c]{@{}l@{}}\\\\Paige had 8 songs on her mp3 \textcolor{red}{players ,} \\If she deleted 5 \textcolor{red}{previous} songs from it and \\then added 30 new songs, how many \\songs does she have on her mp3 player ?\\\\\end{tabular} \\ \hline
\begin{tabular}[c]{@{}l@{}}\\\\A \textcolor{green}{plane} travels 1200 miles in 3 hours . At \\ the same rate , how many additional hours \\ would it take to travel an additional 2000 \\ miles ?\\\\\end{tabular} &
  \begin{tabular}[c]{@{}l@{}}\\\\A \textcolor{red}{helicopter} travels 1200 miles in 3 hours . \\At the same rate , how many additional \\hours would it take to travel an additional \\2000 miles ?\\\\\end{tabular} \\ \hline
\begin{tabular}[c]{@{}l@{}}\\\\\textcolor{green}{Farmer} brown ' s farm is 200 acres , and \\ farmer smith ' s farm is 100 acres more \\ than twice that . how many acres do the \\ two farms have , together ?\\\\\end{tabular} &
  \begin{tabular}[c]{@{}l@{}}\\\\\textcolor{red}{.} brown ' s farm is 200 acres , and \\ farmer smith ' s farm is 100 acres more \\ than twice that . how many acres do the \\ two farms have , together ?\\\\\end{tabular} \\ \hline
\begin{tabular}[c]{@{}l@{}}\\\\Jack \textcolor{green}{had} \$ 100 . Sophia gave him 1 / 5 \\ of her \$ 100 . how many dollars does \\ Jack have now ?\\\\\end{tabular} &
  \begin{tabular}[c]{@{}l@{}}\\\\Jack \textcolor{red}{got} \$ 100 . Sophia gave him 1 / 5 \\ of her \$ 100 . how many dollars does \\ Jack have now ?\\\\\end{tabular} \\ \hline
\end{tabular}
}
\caption{Samples of RobustMath (right) and their original samples (left) .}
\label{robustM}
\end{table*}

\end{document}